\documentclass[10pt,twocolumn,letterpaper]{article}

\usepackage{cvpr}
\usepackage{times}
\usepackage{epsfig}
\usepackage{graphicx}
\usepackage{amsmath}
\usepackage{amssymb}
\usepackage{booktabs}
\usepackage{url}            
\usepackage{booktabs}       
\usepackage{amsfonts}       
\usepackage{nicefrac}       
\usepackage{microtype}      
\usepackage{graphicx}

\usepackage[breaklinks=true,bookmarks=false]{hyperref}

 \cvprfinalcopy 

\def\cvprPaperID{****} 

\ifcvprfinal\fi
\begin{document}

\title{Coherent and Controllable Outfit Generation}

\author{Kedan Li, Chen Liu, and David Forsyth\\
University of Illinois at Urbana-Champaign\\
{\tt\small \{kedanli2, chenliu8, daf\}@illinois.edu}
}

\maketitle

\begin{abstract}

When thinking about dressing oneself, people often have a theme in mind whether they're going to a tropical getaway or wish to appear attractive at a cocktail party. A useful outfit generation system should come up with clothing items that are compatible while matching a theme specified by the user. Existing methods use item-wise compatibility between products but lack an effective way to enforce a global constraint (e.g., style, occasion). 

We introduce a method that generates outfits whose items match a theme described by a text query. Our method uses text and image embeddings to represent fashion items. We learn a multimodal embedding where the image representation for an item is close to its text representation, and use this embedding to measure item-query coherence. We then use a discriminator to compute compatibility between fashion items. This strategy yields a compatibility prediction method that meets or exceeds the state of the art.

Our method combines item-item compatibility and item-query coherence to construct an outfit whose items are (a) close to the query and (b) compatible with one another. Quantitative evaluation shows that the items in our outfits are tightly clustered compared to standard outfits. Furthermore, outfits produced by similar queries are close to one another, and outfits produced by very different queries are far apart. Qualitative evaluation shows that our method responds well to queries. A user study suggests that people understand the match between the queries and the outfits produced by our method.

%


\end{abstract}

\begin{figure}
  \centering
  \includegraphics[width=1\linewidth]{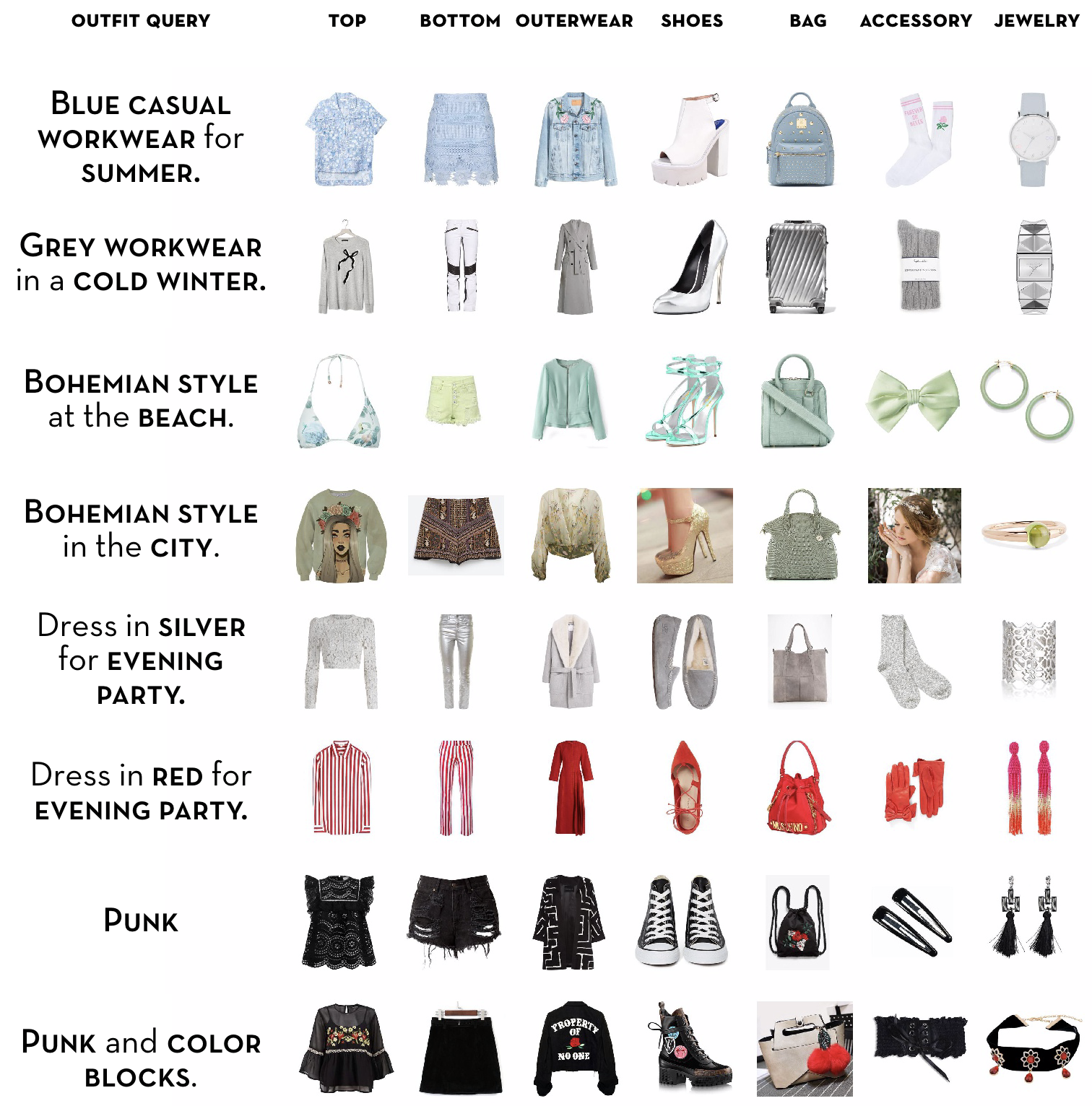}
  \caption{ 
Our method generates outfits that match well with the query sentence describing the outfit. When the modifier (color, seasons, location, style, etc.) of the query changes, the resulting outfit changes accordingly. For example, when we query for ``bohemian style" outfit, summer clothing with decorations shows up. And when we add ``city" and ``beach" as different modifiers, the ``city" modifier results in a darker tone and chunky shoes, while the ``beach" modifier gets bikini and mini pants.}
  \label{fig:example_query}
\end{figure}

\section{Introduction}

Outfit generation is a high-value task for both individuals and the e-commerce industry. In reality, people often build outfits centered around a theme, which could be a style, an event or a season, etc. A recent study suggests that ``how to dress for occasions, activities, and seasons" is one of the most common questions users ask stylists online~\cite{Vaccaro:2018:DFP:3173574.3174201}. While people can often accurately describe their requirements for an outfit using short sentences, composing outfits from a large pool of items is a challenging and time-consuming task, even for professionals.

A large body of prior work shows success in measuring compatibility between clothing items and composing outfits~\cite{Veit2015, He2016LearningCA, DBLP:journals/corr/abs-1803-09196, Cucurull2019ContextAwareVC, Hsiao2017CreatingCW}. But these outfit generation procedures heavily rely on item-item compatibility --- whether all items in an outfit are compatible with one another. However, a set of pairwise compatible fashion items is not necessarily an outfit --- the fashion items may not form a consistent impression. Without proper measurement, their methods cannot enforce a theme in an outfit level. Other works learn multimodal embeddings of fashion items using both image and text~\cite{7780408, fashion-search, fashion-concept-disco, Hsiao2017LearningTL, 7298688}. Using product descriptions as a similarity metric allows the embedding to capture meaningful information such as styles, attributes, and support robust product retrieval. In this work, we combine the merit of both groups of prior work to enable outfit generation guided by a description.

To compose outfits coherent with a chosen sentence, our model must learn to measure both the compatibility between two fashion items and the coherence between items and outfit descriptions (item-query coherence). We extract features for fashion items using a multimodal item encoder and predict the compatibility between two items using discriminator networks. From the fashion item features, we also learn an image-text embedding to measure the coherence between items and sentences describing outfits. Because the product descriptions often discuss high-level styles and context, the learned embedding can effectively capture outfit level themes. Finally, we describe an outfit generation method that selects fashion items one at a time using pairwise compatibility and item-query coherence measured by the model.

We evaluate our method's performance on both item-wise compatibility and item-query coherence using published outfit datasets. Results show that our item-wise compatibility prediction matches or exceeds the state of the art methods on established metrics. Quantitative evaluation for outfit generation procedure is difficult, and we are not aware of any prior work that attempted. Thus, we introduce a new measurement. We regard outfits as clusters in our embedding space and show that outfits generated using our method have smaller cluster sizes (is more coherent) than baseline outfits. Distances between cluster centers also positively correlate with the distances between query sentences. Qualitative examples demonstrate the fine-grain level of control our method enables. For example, the query ``bohemian in the city" and ``bohemian at the beach" both produce outfits matching the bohemian style and yet also adapting to each modifier appropriately (Figure~\ref{fig:example_query}). Through a preliminary user study, we demonstrate that outfits generated by our method are coherent.

Our contributions are: 

\begin{itemize}
    \item We are the first to introduce a method that can generate outfits guided by a sentence describing the outfit.
    \item Our method generates outfits that are coherent and consistent with the query, demonstrated through quantitative and qualitative evaluation.
    \item Our work achieves or exceeds state of the art performance on compatibility prediction on established dataset and metrics.
\end{itemize} 

\section{Related Work}
\subsection{Item-wise Fashion Compatibility} 
    \label{related_work_compat}

Recent works learn to model the compatibility relationship between clothing items using product images from annotated outfits. 

\paragraph{Pairwise compatibility} Some works formulate the fashion compatibility problem as predicting a pairwise relationship and learn embeddings using compatibility as the distance metric~\cite{McAuley:2015:IRS:2766462.2767755, Veit2015}. Vasileva et al. achieved improvement in performance by projecting image features from a similarity embedding to type-aware compatibility embeddings and using a negative sampling method that also respects types~\cite{DBLP:journals/corr/abs-1803-09196}. Song et al. utilized the fashion domain knowledge to learn the compatibility in a teacher-student schema~\cite{Song:2018:NCM:3209978.3209996}. 

\paragraph{Multi-way compatibility} Other methods model an outfit as a whole --- combine the representation of many items. For example, Han et al. learned a visual-semantic embedding by training a bidirectional LSTM model to predict the next compatible item and generate an outfit sequentially~\cite{DBLP:journals/corr/HanWJD17}. Recently, Cucurull et al. used graph convolutional network to model compatibility by regarding items as nodes and compatibility relationship as edges, and trained the model to predict missing edges~\cite{cucurull2019context}. The method outperformed state of the art on compatibility prediction by aggregating information from known-compatible items.

In summary, multi-way compatibility performs better with a partially-complete outfit, while pairwise compatibility prevails without that extra information.

\subsection{Multimodal fashion embedding}
    \label{related_joint_embedding}

Both image and text describe fashion items. Images provide rich visual features such as color, shape, and texture, while text information contains categorical, contextual, and functional attributes that may be difficult to determine visually. Recent works learned joint embeddings between fashion item images and text to capture the stylistic and contextual relationship between fashion items. Simo-Serra et al. demonstrated that text information like tags could provide weak labels for feature extraction~\cite{7780408, 7298688}. Han et al. trained joint visual semantic embeddings with product images and descriptions in the form of word vectors to retrieve spatially-aware fashion concepts~\cite{fashion-concept-disco}. Hsiao et al. extracted style using a topic model and learned an image-topic joint embedding to enable style retrieval and measuring style-coherence~\cite{Hsiao2017LearningTL}. Yang et al. trained a compatibility embedding using a multimodal item encoder and showed that both image and text features contribute to predicting compatibility~\cite{DBLP:conf/aaai/YangMLWC19}.

\begin{figure*}
\begin{center}
  \includegraphics[width=.99\linewidth]{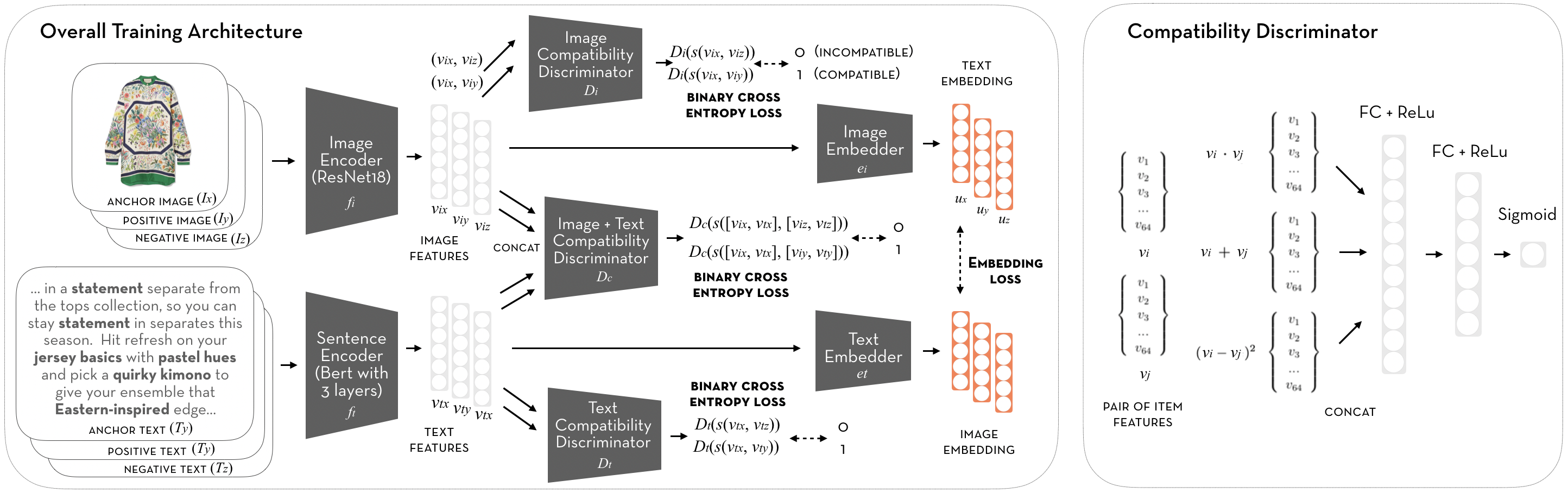}
\end{center}
  \caption{The left diagram shows the overall architecture of our method during training. The architecture consists of a multimodal item encoder, an image-text embedding, and three discriminators. The embedding measures the coherence between fashion items and the outfit query. The discriminators predict compatibility from pairs of image, text, or both features concatenated. The right diagram shows the detail implementation of a discriminator. It computes the sum, the dot product and the squared difference of the two items' features, and pass through a fully connected network with two hidden layers with ReLu nonlinearity.}
  \label{fig:model_diagram}
\end{figure*}

\subsection{Outfit Generation}

Previous outfit generation procedures mostly rely on item-wise compatibility measurements~\cite{DBLP:journals/corr/abs-1803-09196, Hsiao2017LearningTL, Bettaney2019FashionOG, 3206025.3206048}. For example, the BiLSTM method allowed input text only to influence the selection of the starting item, and generate the rest of the outfit based on item-wise compatibility~\cite{DBLP:journals/corr/HanWJD17}. Nakamura et al. extended the BiLSTM method to provide additional guidance on outfit generation using a style-vector obtained from an auto-encoder~\cite{Nakamura2018OutfitGA}. However, the work did not include a meaningful evaluation. Using compatibility metric, Hsiao et al. introduce a method for generating capsule wardrobe --- a minimal set of items that generate a maximal number of compatible outfits~\cite{Hsiao2017LearningTL}. Feng et al. proposed a system for evaluating outfit composition through interpretable attributes matching, but did not allow explicit input from users ~\cite{3206025.3206048}. Chen et al. combined user preferences with compatibility metrics to incorporate personalization into the outfit generation process~\cite{Chen:2019:PPO:3292500.3330652}.

In summary, previous outfit generation methods emphasize the compatibility relationship between items in an outfit. However, it is important to notice that a set of pairwise compatible products do not always form a coherent outfit. Items in an outfit should work together to create an overall impression. By incorporating the prior work in Section~\ref{related_work_compat} and ~\ref{related_joint_embedding}, we introduce a new method to generate outfits with fine-grain control over how the items in the outfit cohere.

\section{Methodology}

We wish to control the outfit generation process using a text description of the outfit. To achieve this, we need to measure (a) how coherent items are with a written description and (b) how compatible items are with one another. As our fashion items come with images and text descriptions, we use both images and text to produce an embedding. This embedding is used to measure the coherence between items and the outfit description. Because coherence is a different notion from item-wise compatibility, we cannot reuse embedding distance directly to measure item-wise compatibility, but instead, use a discriminator. The image-text embedding and the compatibility discriminators are trained jointly and share the same item feature encoder, as shown in Figure~\ref{fig:model_diagram}. Using the trained model, we describe an outfit generation algorithm that composes outfits by selecting one item per step. At each step, the algorithm uses the metric learned by the model to choose the next fashion item for the outfit. 

\subsection{Multimodal Item Encoder}

Our method encodes both image and text information of a fashion item $X = (I_x, T_x)$ into two feature vectors $v_i, v_t$. The image encoder $v_i = {f_i}({I_x})$ uses ResNet18 initialized with ImageNet pre-trained weights~\cite{He2015DeepRL}. The text encoder $v_t = {f_t}({T_x})$ uses BERT with three hidden layers initialized with pre-trained weights provided by Google~\cite{devlin2018bert}. We fine-tune the weights of the item encoders when training the embedding and the compatibility discriminators. Our multimodal item encoder architecture is similar to that of Yang et al. but uses different backbone architectures~\cite{DBLP:conf/aaai/YangMLWC19}.

\subsection{Multimodal Embedding}

We learn an image-text embedding to measure how closely fashion items are related to outfit descriptions. Since we are not aware of any dataset that provides outfit level descriptions of high quality, we use product descriptions as the metric for learning the embedding. Since product descriptions often discuss styles and contexts that match the product (as the product description snippet in Figure~\ref{fig:model_diagram}), the learned embedding also captures outfit-level themes. To obtain an image-text embedding, we learn an image embedder $e_i({v_i})$ that transforms image feature vector into embedding vector $u_{ix}$. We also learn a text embedder $e_t({v_t})$ that maps text feature vector into embedding vector $u_{tx}$. We encourage the euclidean distance between $u_{ix}$ and $u_{tx}$ to be smaller than the distance between them and a third, different item $Y$ (represented as $u_{iy}, u_{ty}$) by minimizing the triplet loss~\cite{7298682}
\[
\mathcal{L}_m = {\mathcal{L}}_{t}(u_{ix}, u_{tx}, u_{ty}) + {\mathcal{L}}_{t}(u_{ix}, u_{tx}, u_{iy}) 
\]
where
\[
\mathcal{L}_{t}(a, p, n)= max( \|f(a) - f(p)\|^2 - \|f(a) - f(n)\|^2 + \alpha, 0)
\]. Additionally, we minimize the squared distance between $u_i$ and $u_t$ and impose an $\l_2$ regularization on the linear layer producing the embedding.\[
\mathcal{L} = {\mathcal{L}}_{m} + \|u_t - u_i\|^2 + {\mathcal{L}}_{2}
\] 
The encoder functions $e_i$ and $e_t$ consist of a fully connected neural network of 1 hidden layer with ReLU non-linearity. 

\subsection{Compatibility Discriminator}

The discriminator learns a non-linear function $D$ that takes a pair of fashion item features $({v_x}, {v_y})$ as input, and produces a compatibility score in the range $[0, 1]$. This score will be tested against a threshold, to be chosen later. From the item features, we compute a joint feature vector ${s}({v_x}, {v_y})$, then apply a fully connected network to obtain the compatibility score. We train three different discriminators using image features $v_{ix}$, text features $v_{tx}$, and both features concatenated $[v_{ix}, v_{tx}]$ where
\[
D_i({I}_x, {I}_y)=D_i({s}(v_{ix}, v_{iy}))
\]
\[
D_t({T}_x, {T}_y)=D_i({s}(v_{tx}, v_{ty}))
\]
\[
D_c({I}_x, {T}_x, {I}_y, {T}_y)=D_c({s}([v_{ix}, v_{tx}], [v_{iy}, v_{ty}])))
\]
Training data consists of pairs that are compatible or incompatible, and the discriminator $D$ is learned using binary cross-entropy loss. The discriminator $D$ consists of a fully connected neural network with 2 hidden layers with ReLU non-linearity and an output layer with sigmoid activation. Although the concatenated discriminator largely outperforms the others (demonstrated in section~\ref{compatibility-results}), it is useful to have the image discriminator and the text discriminator in case one source of information is not available.

The function $s$ transforms a pair of item representations into a relationship representation. We investigate various choices of ${s}$, constructed to be symmetric in their arguments.
We consider the following options: {\bf dot}, where ${s}_{dot}({a}, {b}) = \left[a_1 b_1, \ldots, a_d b_d\right]^T$; {\bf diff}, where ${s}_{diff}({a}, {b}) = \left[(a_1-b_1)^2, \ldots, (a_d-b_d)^2\right]^T$; {\bf sum}, where ${s}_{sum}({a}, {b}) = {a}+{b}$.  
Through experiments, we found that supplying all three to the discriminator function yields better results than each of them individually. We write the concatenation of the feature vectors of
\textbf{dot}$+$\textbf{sum}$+$\textbf{diff} as $\left[{s}_{dot}^T, {s}_{sum}^T, {s}_{diff}^T\right]^T$.

\begin{figure}
\begin{center}
  \includegraphics[width=.99\linewidth]{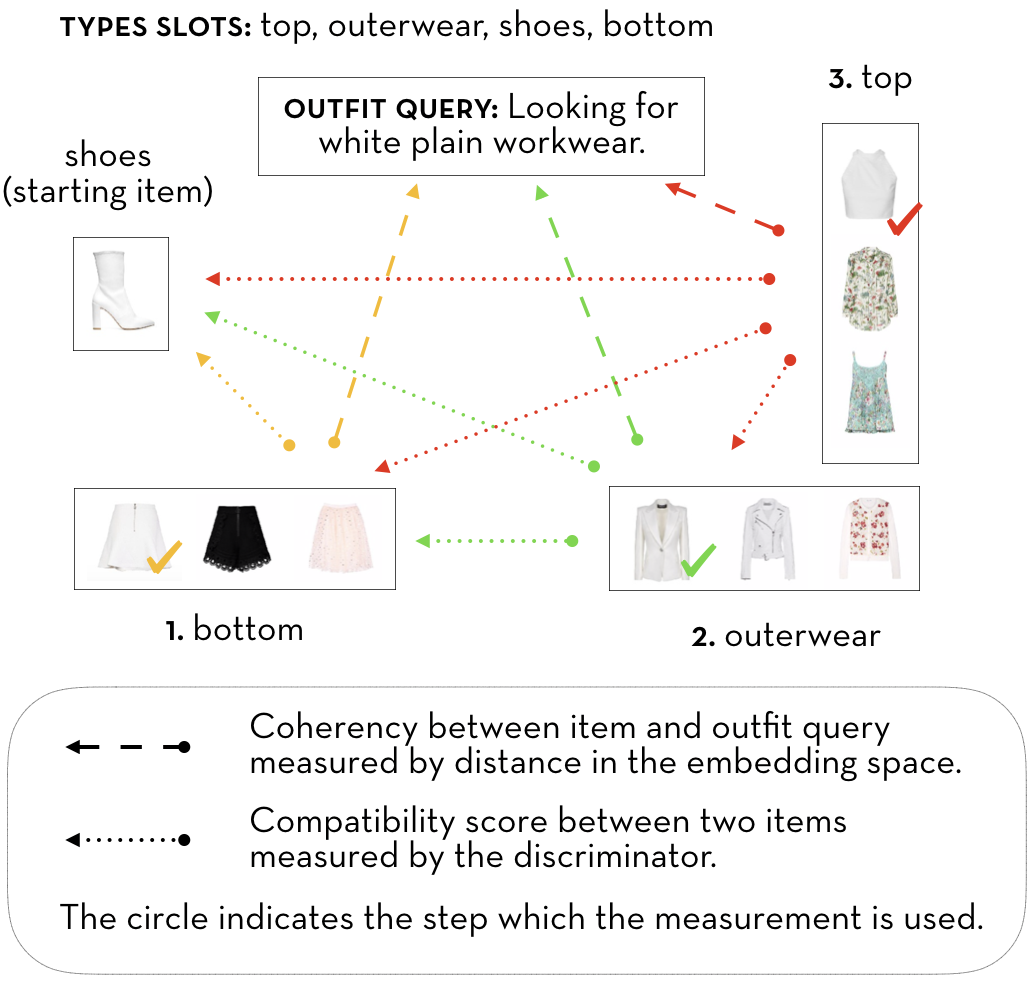}
\end{center}
    \label{fig:outfit_generation_procedure}
  \caption{The diagram shows our iterative outfit generation procedure. We rank all candidate items of the chosen type based on coherence and item-item compatibility measured by the model. We then add one of the top-ranked items to the outfit and repeat the process with the next missing type until all the type-slots are filled.}
\end{figure}

\paragraph{Generating the training data}
    \label{generate_training_data}
All pairs of items that occur in an outfit are assumed compatible. Therefore, we use all such pairs in the training outfits as positive examples. As is usual for embedding methods, we assume that an arbitrary pair that does not appear in any of the training outfits is incompatible. When randomly sampling negative items, we adopt the type-aware negative sampling method introduced by~\cite{DBLP:journals/corr/abs-1803-09196}, which requires the negative item to have the same product category as the positive items.

\subsection{Outfit Generation Procedure}

We introduce an iterative outfit generation algorithm which completes a partial outfit one item at a time, as shown in Figure~\ref{fig:outfit_generation_procedure}. We define a partial outfit as a list of desired item types (e.g., tops, bottoms) and some optional starting item(s). At each step, the algorithm first selects a missing item type whose items have the smallest average distance to the outfit query $q$. The algorithm then computes a rank for all candidate items $X$ of the chosen type. (a) If the outfit has no existing item, the algorithm ranks the items by their distances with the outfit query text $r_x=\|q - u_x\|$, measured in the embedding space. (b) If some item(s) $Y$ exists in the outfit, the model computes the average compatibility score between every item in the outfit and each candidate item $c_x = (\sum_{y \in Y} D(s({v_{x}, v_{y}})))/|Y|$, and rank the candidate items based on $r_x=\|q - u_x\|/{c_x}$. Then, the algorithm adds one of the top-ranked items to the outfit and repeats with the next missing type until all empty slots are filled.

For selecting an item from the ranked list at each step, we ask the user to pick a threshold $k$ and sample an item uniformly and at random from the top $k$ items. Using $k=1$ results in the optimal outfit but lacks diversity. Using a large $k$ increases the diversity of the generated outfits at the cost of outfit quality. An improved sampling method uses a bias distribution where the probability an item $i$ get sampled is $e^{-r_i}/(\sum^k_{j=1} e^{-r_j})$. Sampling from a bias distribution effectively increases the probability of selecting top-ranked items while preserving some randomness.

\section{Experimental Setup}


\subsection{Dataset}
\label{section:dataset}

When modeling fashion compatibility, curated outfits are a common source of supervision signals and the ground truth for evaluation. Many public outfit datasets (FashionCV by Song et al., Maryland Polyvore by Han et al., Polyvore Outfits, and Polyvore Outfits Disjoint by Vasileva et al.) consist of outfits mined from Polyvore~\cite{Song:2017, DBLP:journals/corr/HanWJD17, DBLP:journals/corr/abs-1803-09196}. Polyvore is a social network where fashion lovers curate outfits using a set of fashion product images. Other fashion compatibility datasets, such as the Amazon products dataset, consist of co-purchase relationships between products but are not suitable for outfit generation task~\cite{McAuley:2015:IRS:2766462.2767755}. For the multimodal embedding to encode contextual information, our model requires the dataset to have rich descriptions attached to the fashion items. However, FashionCV and Maryland Polyvore dataset only provide fashion product title and category, which are insufficient for embedding contextual information. 

{\bf Polyvore Outfit} consists of 68,306 unique outfits and 365,054 unique items. Each product item has one image, and the metadata includes title, description, and product category. {\bf Polyvore Outfits Disjoint(-D)} consists of 32,140 unique outfits and 175,485 unique items and is a subset of Polyvore Outfits. Polyvore Outfits Disjoint dataset is also the only public outfit dataset in which product items do not overlap between splits (training, validation, and testing). To prevent data leakage, we only use {\bf Polyvore Outfits-D} to evaluate item-query coherence, generate qualitative examples, and produce examples for our user study. 

\subsection{Training details} 
\label{section:training}

We train the multimodal embedding and the three compatibility discriminators jointly. Each training runs for 20 epochs using Adam optimizers with a learning rate of 1e-5, beta1 of 0.9, and beta2 of 0.999~\cite{Kingma2014AdamAM}. Each epoch runs over every positive (compatible) pair once. At every epoch, we resample negative pairs using the sampling method in Section~\ref{generate_training_data}. Due to the imbalance nature of the learning problem, the model often converges at the fourth or the fifth epoch. The image features, text features, and multimodal embedding are 128-dimensional unless otherwise specified. 

\section{Compatibility Experiments}

In this section, we establish that our compatibility prediction performance matches or outperforms state of the art using the datasets described in Section~\ref{section:dataset}.

\subsection{Metrics} Following prior work, we evaluate our method on the fashion compatibility task and the fill-in-the-blank (FITB) task. Fill in the blank measures a method's accuracy in choosing the correct item to fill in a blank in an outfit, from a list of four items~\cite{DBLP:journals/corr/HanWJD17}. For fashion compatibility tasks, we follow the usual practice of averaging the score for all pairs of items in an outfit, then computing AUC for ground truth outfits against random outfits. For both tasks, we use the identical test set produced by Han et al. on the Maryland Polyvore dataset and Vasileva et al. on Polyvore Outfits dataset~\cite{DBLP:journals/corr/HanWJD17, DBLP:journals/corr/abs-1803-09196}. 

\subsection{Compared Approaches} 

\begin{itemize}
    \item {\bf Siamese Net} (Veit et al.~\cite{Veit2015}) estimates the pairwise compatibility with the euclidean distance.
    \item {\bf BiLSTM} (Han et al.~\cite{DBLP:journals/corr/HanWJD17}) is trained to predict the next fashion item sequentially conditioned on exiting items. 
    \item {\bf CSN} (Vasileva et al.~\cite{DBLP:journals/corr/abs-1803-09196}) models the compatibility with conditional similarity networks to produce type-aware visual embeddings. 
    \item {\bf GCN} (Cucurull et al.~\cite{Cucurull2019ContextAwareVC}) learns a graph convolutional network to predict the compatibility conditioned on adjacent items. 
\end{itemize} 

Compared approaches are trained using their provided code with the default parameters unless specified. For a fair comparison with the graph convolutional network, we provide it with zero or one ($k=0$ or $k=1$) additional known compatible item within the outfit and a maximum of three known compatible items out of the outfit at test time. We adopt this setting because the compatibility prediction must perform well at the beginning of the outfit generation process --- when the outfit only has one or a few items.

\begin{table}
  \caption{
The table compares the compatibility prediction performance of our method to state of the art on Maryland Polyvore, Polyvore Outfits (Polyvore), and Polyvore Outfits Disjoint (Polyvore-D) datasets. For our method, we compare the performance of using only image features, only text features, and concatenating image and text features (Cat).
}
  \label{methods-comparison-table}
  \centering
  \scalebox{0.9}{
  \begin{tabular}{l|llllll}
    \toprule
     & \multicolumn{2}{c}{Maryland} & \multicolumn{2}{c}{Polyvore} & \multicolumn{2}{c}{Polyvore-D}\\
   	Method & AUC & FITB & AUC & FITB & AUC & FITB \\
    \midrule
    BiLSTM & .901 & 68.6 & .658 & 40.2 & .621 & 39.6 \\
    Siamese & .853 & 53.7& .814 & 53.4 & .812 & 52.2 \\    
    CSN (128D) & .951 & 86.0 & .864 & 55.3 & .835 & 54.1 \\
    CSN (512D) & \textbf{.956} & \textbf{87.2} & .814 & 53.4 & .812 & 52.2 \\
    GCN (k=0) & .698 & 59.5 & .677 & 51.3  & .669 & 49.5\\
    GCN (k=1) & .849 & 66.7 & .817 & 63.0 & .865 & 62.6\\
        \cmidrule(r){1-7}
    Ours (Text) & .850 & 54.8 & .832 & 51.0 & .780 & 45.7\\
    Ours (Image) & .891 & 59.8 & .895 & 59.1 & .852 & 55.9 \\
   	Ours (Cat) & .931 & 69.9 & \textbf{.928} & \textbf{66.1} & \textbf{.901} & \textbf{62.8} \\
    \bottomrule
  \end{tabular}
  }
\end{table}

 \begin{table}
  \caption{
The table compares the performance of the compatibility discriminators using different transformations $s$ on the Polyvore Outfits dataset. All conditions only use image features.
  }
  
  \label{ablation-table}
  \centering
  \begin{tabular}{l|lll}
    \toprule
   	Transformation Function & AUC & FITB \\
    \midrule
    diff & .853  & 52.5\% \\
    sum & .858  & 53.5\% \\
    dot & .878  & 56.1\% \\
    diff $+$ sum & .879 & 56.8\% \\
    dot $+$ diff & .889 & 57.6\% \\
   	dot $+$ sum & .892 & 58.6\% \\
   	dot $+$ diff $+$ sum & \textbf{.895} & \textbf{59.1\%} \\
    \bottomrule
  \end{tabular}
\end{table}

\subsection{Results}
\label{compatibility-results}

As shown in Table~\ref{methods-comparison-table}, our method outperforms prior work on the Polyvore Outfits dataset but performs less well on the Maryland Polyvore dataset. We believe the performance drop is caused by the absent of product descriptions in the Maryland Polyvore dataset. When using image information alone, the method achieves similar performance with prior work, and combining text information yields an average of 5\% improvements. Using text information alone yields less accuracy, as expected.

Table~\ref{ablation-table} shows the comparison between different feature combinations of the transformation functions. Using transformation of \textbf{dot} alone performs 2\% better than \textbf{diff} and \textbf{sum} alone. Concatenating either \textbf{diff} or \textbf{sum} with \textbf{dot} increase the performance by 1\%. Concatenating the three relationships achieves the best performance with an additional improvement of 1\%.

\section{Generating Coherent Outfit}

\begin{figure*}
\begin{center}
  \includegraphics[width=.99\linewidth]{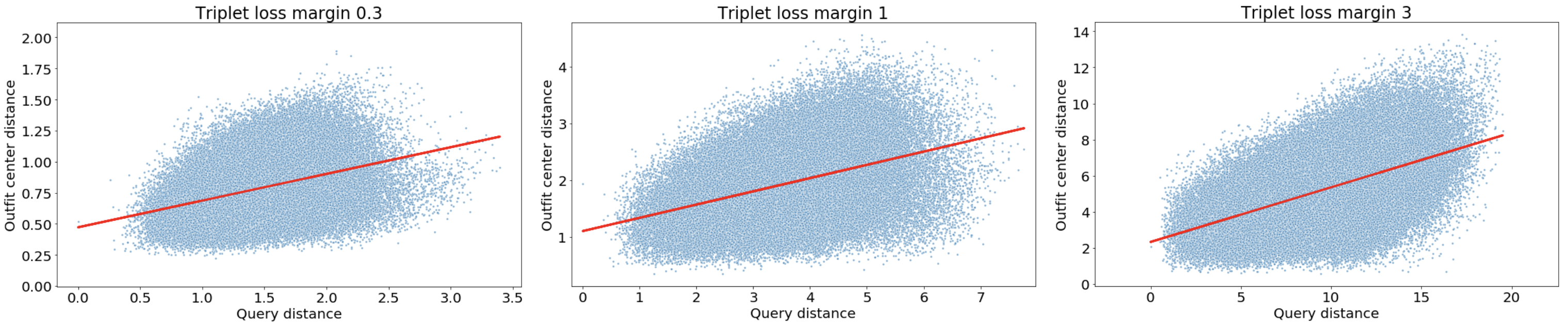}
\end{center}
  \caption{The figure shows a scatter plot between outfit query distances and outfit center distances of all possible pairs in 500 generated outfits. The X-axis shows outfit query distances and the y-axis shows outfit center distances. Both axis are in the same scale. The plot shows a positive correlation between the two distances and the correlation coefficient increases as the margin of the embedding loss increases. }
  \label{fig:scatter_plot}
\end{figure*}

\begin{figure*}
\begin{center}
  \includegraphics[width=.99\linewidth]{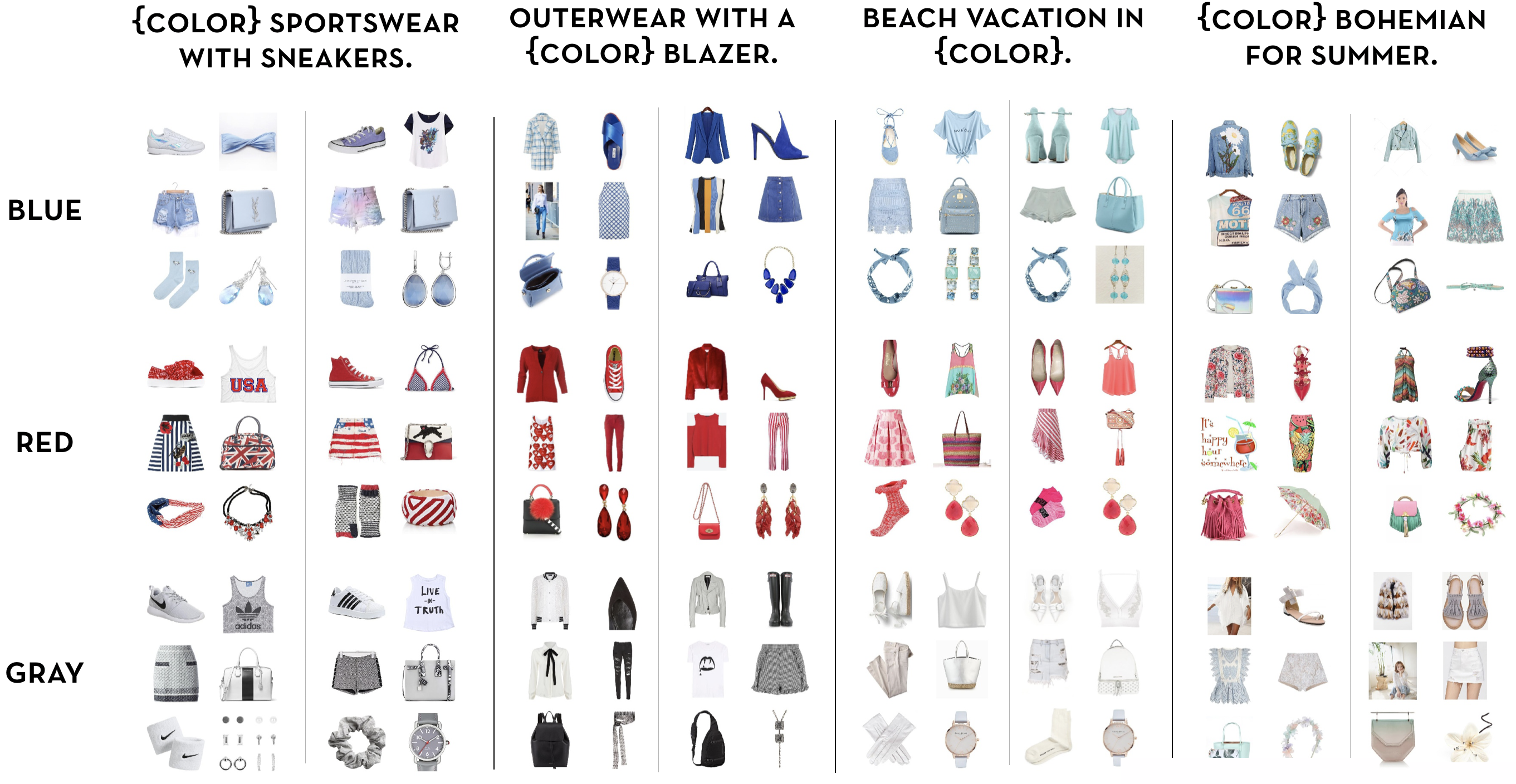}
\end{center}
  \caption{
The diagram shows the outfits produced from different outfit queries (horizontal axis) paired with varying modifiers of color (vertical axis). Each cell has two outfits. The outfits respond to both specific items and abstract concepts (style, occasion, seasons) --- when asking for sportswear with sneakers, sneakers show up; when asking for bohemian for summer, items match the style and season. Outfits in each row also have very consistent color, suggesting that the model respond well to color modifiers.
   }
  \label{fig:color_filled_query}
\end{figure*}

\begin{figure}
\begin{center}
  \includegraphics[width=.99\linewidth]{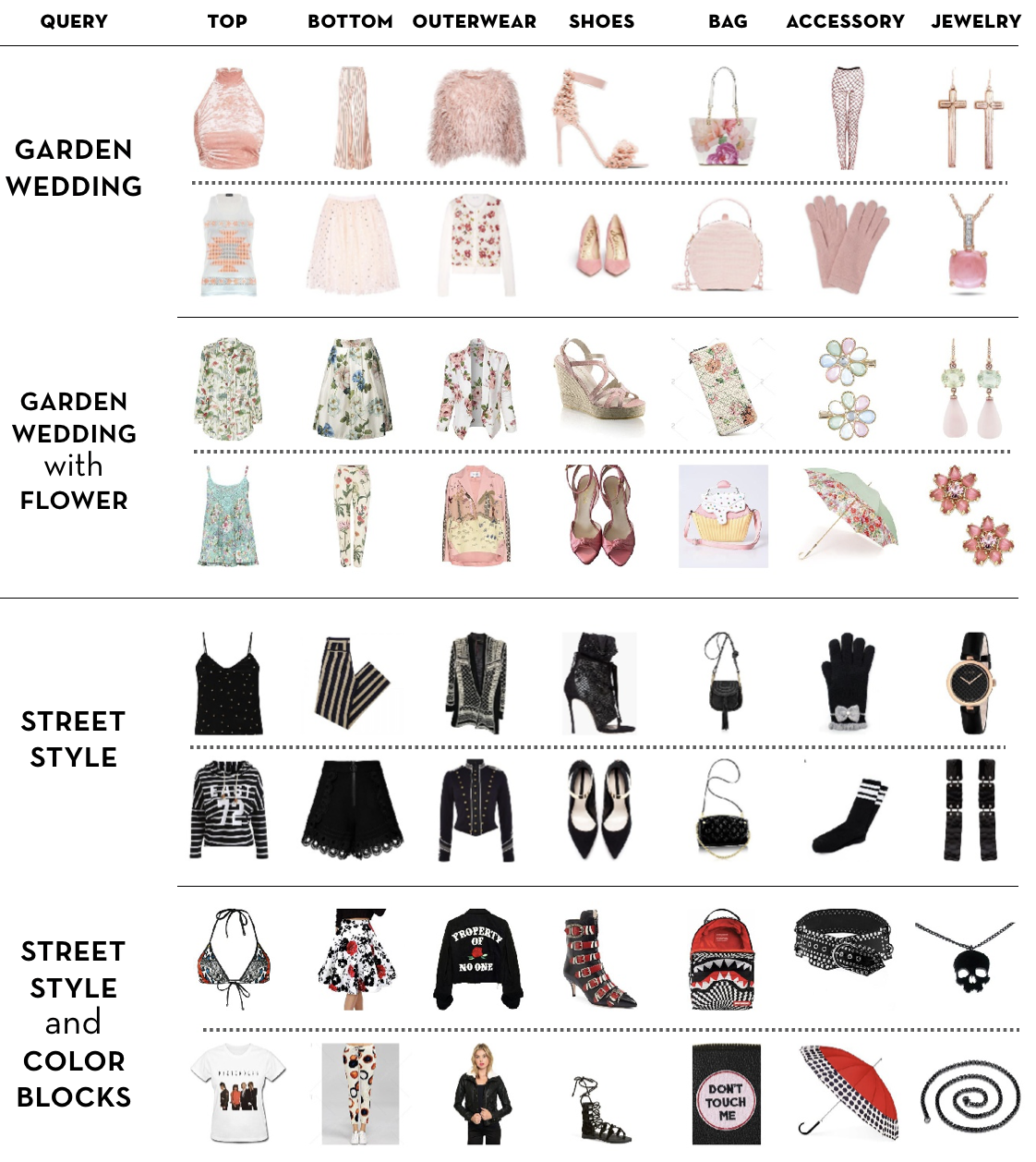}
\end{center}
  \caption{The diagram compares queries with and without modifiers. Each row is an outfit; each query has two outfits. The generated outfits correctly respond to modifiers. For example, when we attach ``with flower" to the ``garden wedding" query, floral patterns show up in the resulting outfit.
}
  \label{fig:comparison_query}
\end{figure}

\begin{figure}
\begin{center}
  \includegraphics[width=.99\linewidth]{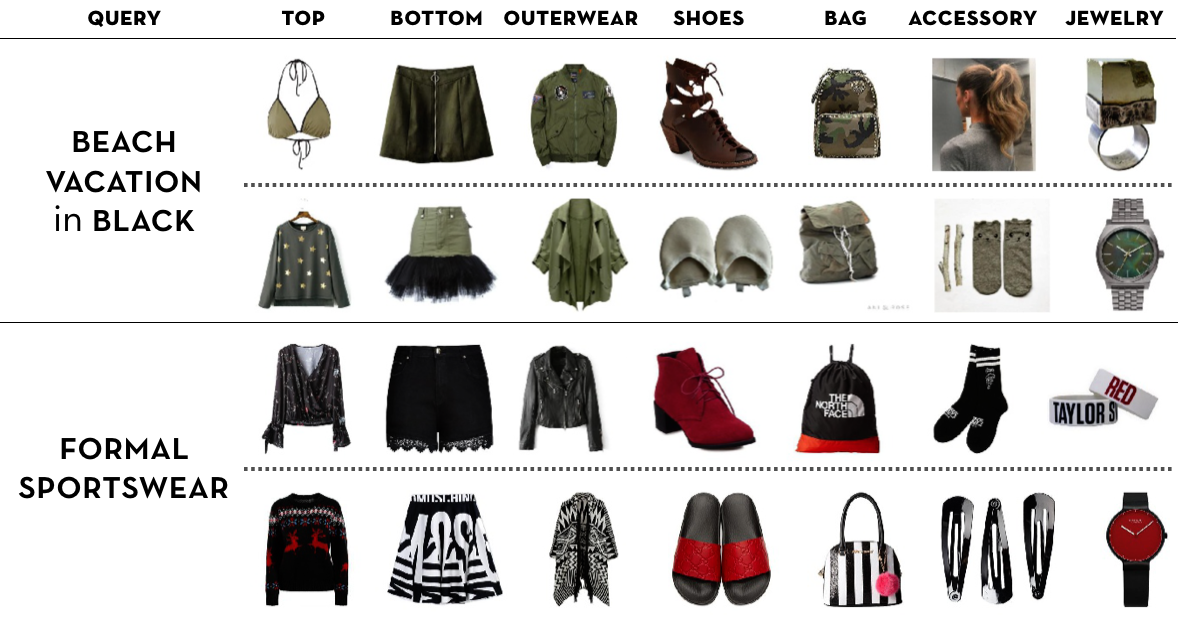}
\end{center}
  \caption{This diagram show results from queries that are difficult or self-contradictory. Each row is an outfit; each query has two outfits. With an unnatural query, we see odd results as expected. 
}
  \label{fig:bad_queries}
\end{figure}

 \begin{table}
  \caption{
 The left columns show the average size ($s_b$) of 500 outfits generated using a baseline method (CSN) and the average size ($s_c$) of 500 outfits created using our method. The middle columns show the average distances ($d_q$) between outfit query and the average distance ($d_c$) between outfit centers. The average is computed among all pairs of generated outfits. The right columns show the correlation coefficient ($\rho$), p-value ($p$), and R-squared measure ($R^2$) between outfit query distances and outfit center distances. 
  }
  \centering
    \scalebox{0.85}{\begin{tabular}{l|ll|ll|lll}
    \toprule
    	 margin ($\alpha$) & $s_b$ & $s_c$ & $d_q$ & $d_c$ & $\rho$ & $p$ & $R^2$ \\
    \midrule
    $\alpha$ = 0.3 & 2.35 & 2.07 & 1.38 & 0.81 & 0.34 & 0.0 & 0.14 \\
    $\alpha$ = 1 & 5.70 & 4.93 & 3.21 & 2.14 & 0.42 & 0.0 & 0.17 \\
    $\alpha$ = 3 & 16.84 & 13.31 & 8.85 & 6.83 & 0.49 & 0.0 & 0.28\\
    \bottomrule
  \end{tabular}}
   \label{context-table-a}
\end{table}

Quantitative evaluation of an outfit generation procedure is difficult, and we are aware of no prior attempt to do this. In this work, we evaluate our method using two quantitative measurements of coherence between the query sentence and the generated outfit. We also use qualitative examples to show that our approach constructs outfits that are responsive to queries. A user study shows that our outfits were deemed coherent.

\subsection{Quantitative Evaluation}

Items that are nearby in the embedding space share a similar theme. We regard an outfit as a small cluster in the embedding space and ask the following questions:
\begin{itemize}
    \item {\bf Question 1:} are the clusters generated by our method smaller than those generated by a baseline method?
    \item {\bf Question 2:} when two queries are far apart/close, are the centers of the generated outfit clusters far apart/close?
\end{itemize}

\paragraph{Outfit Cluster Size} We define outfit cluster size as the median distance between each item in the outfit and the center of the outfit. We compute the center of an outfit as the mean of all item vectors measured in the embedding space. The center cannot be measured using a medoid because two outfits with shared item(s) may result in zero distance. A small cluster size suggests that an outfit has high coherence. We expect the average cluster size of outfits generated using our coherent generation method to be smaller than the average cluster size of outfits made by a baseline method that only considers item-wise compatibility.

\paragraph{Cluster Size Comparison}
We generate 500 outfits from our method and 500 outfits from a baseline method, both using products in the test set as candidate items. We use CSN as the baseline generation method, which only considers pairwise compatibility. When generating with our method, we randomly sampled product descriptions from the validation set as queries. When selecting items from a ranked list, we use the threshold $k=10$ and sample from the bias distribution. The average size of the outfits generated from the two methods is shown in Table~\ref{context-table-a}. We observe that the average outfit size generated by the baseline method $s_b$ is consistently larger than the outfit size generated by our method $s_c$. The difference in size is similar to the margin used in the embedding loss. This result shows that our method generates outfits of higher coherence comparing to methods that only use pairwise compatibility.

\paragraph{Query-Outfit Coherence} With a set of outfits generated using different queries, we expect the distance between the queries to positively correlate to the distance between centers of the generated outfits. In other words, if two queries are similar, we expect the resulting outfits to also be close to each other in the embedding space. We measure the Pearson Correlation Coefficient between query distances and outfit center distances for all possible ($124,750$) pairs among the 500 generated outfits. As shown in Figure~\ref{fig:scatter_plot}, the two distances are positively correlated. The correlation coefficient and the R-squared measures increase when the margin increases. Table~\ref{context-table-a} shows the average outfit query distances ($d_q$) and the average outfit center distances ($d_c$). We observe that the ratio between average outfit cluster size and average outfit center distances $s_c/d_c$ decreases when the margin increases. When $s_c/d_c$ is smaller, outfits generated by two distinct queries are less likely to have overlapping items, and thus is more coherent.


\subsection{Qualitative Examples}

We present examples of outfits generated using queries combined with different color words, as shown in Figure~\ref{fig:color_filled_query}. The two queries on the left mention a specific item, while the right two queries use abstract concepts. The resulting outfits contain the fashion item specified in the query and match the abstract fashion concepts (style, occasion, seasons). In each row, the resulting outfits also consistently reflect the color descriptor. However, when the color modifier only describes an item, it nevertheless affects the entire outfit.

To show the fine-grain level of control our system enables, we experiment adding modifiers to outfit queries. In Figure~\ref{fig:comparison_query}, the outfits respond well to modifiers. For example, when we ask for ``street style and color blocks", colorful decorations show up in the result, which sets itself apart from the default monotone street style.

We also experiment with queries that are rare or self-contradictory, with results shown in Figure~\ref{fig:bad_queries}. For example, when querying with ``formal sportswear", the resulting outfit consists of a mix of formal and sporty items but does not cohere, as expected.

More qualitative examples are available in the supplementary materials.

\subsection{User Study}
We perform an offline user study to further evaluate the quality of generated outfits. Our preliminary user study strongly suggests that outfits generated by our method are understandable to people.  We showed each user (29 users in total) a query along with four outfits and asked the user to select the one outfit which corresponds to the query. The 3 distractors were: (a) a baseline outfit without context; (b) a change in, addition of or deletion of a color modifier (for example, "{\bf Blue} casual vacation" to "{\bf Grey} casual vacation"); (c) a change in the query body but keeping the same modifier (for example,  "Blue {\bf casual vacation}" to "Blue {\bf workwear with a sneaker}"). As shown in supplementary materials, those distractors are strong enough to confuse. From the results, the correct outfit was selected in $62.4\%$ out of all questions answered. For all questions, the majority of the answers are correct.

\section{Conclusion}
In this paper, we introduce a method for generating coherent and compatible outfits constrained by short sentences. Experiments show that our model meets and exceeds state of the art in compatibility prediction. Through quantitative and qualitative evaluation, we demonstrate that the proposed outfit generation method accurately responds to queries and generates outfits that are coherent.

{\small
\bibliographystyle{ieee_fullname}
\bibliography{fashion_outfit_2020}
}

\end{document}